\documentclass[12pt, a4paper]{article}
\usepackage[T1]{fontenc}
\usepackage{tgbonum}
\usepackage{times}
\usepackage{amsmath, amssymb, amsfonts}
\usepackage{graphicx}
\usepackage{epstopdf}
\usepackage{caption, subcaption}
\usepackage[margin=01 in]{geometry}
\usepackage[onehalfspacing]{setspace}
\usepackage{xr-hyper}
\usepackage{hyperref}
\usepackage[affil-it]{authblk}
\numberwithin{equation}{section}
\usepackage{amsthm}

\usepackage{authblk}
\usepackage{url}
\usepackage{cite}
\usepackage[table,xcdraw]{xcolor}
\usepackage{lineno} 
\usepackage{float}
\usepackage{tcolorbox}
\usepackage{hhline}

\usepackage{array}
\newcolumntype{C}[1]{>{\centering\let\newline\\\arraybackslash\hspace{0pt}}m{#1}}
\usepackage{siunitx} 

\providecommand{\keywords}[1]
{
  \small	
  \textbf{\textit{Keywords---}} #1
}

\title{Deep Learning for Disease Outbreak Prediction: A Robust Early Warning Signal for Transcritical Bifurcations}
\author[1]{Reza Miry} 
\author[2,*]{Amit K. Chakraborty}
\author[3,4]{Russell Greiner}
\author[5]{Mark A. Lewis}
\author[2]{Hao Wang}
\author[6]{Tianyu Guan}
\author[1]{Pouria Ramazi}
\affil[1]{Department of Mathematics and Statistics, Brock University, ON, Canada}
\affil[2]{Department of Mathematical and Statistical Sciences, University of Alberta, AB, Canada}
\affil[3]{Department of Computing Science, University of Alberta, AB, Canada}
\affil[4]{Alberta Machine Intelligence Institute, AB, Canada}
\affil[5]{Department of Mathematics and Statistics and Department of Biology, University of Victoria, BC, Canada}
\affil[6]{Department of Mathematics and Statistics, York University, ON, Canada}
\affil[*]{Corresponding author: akchakra@ualberta.ca}
\date{}
\begin{document}

\maketitle


\begin{abstract}
Early Warning Signals (EWSs) are vital for implementing preventive measures before a disease turns into a pandemic. While new diseases exhibit unique behaviors, they often share fundamental characteristics from a dynamical systems perspective. Moreover, measurements during disease outbreaks are often corrupted by different noise sources, posing challenges for Time Series Classification (TSC) tasks. In this study, we address the problem of having a robust EWS for disease outbreak prediction using a best-performing deep learning model in the domain of TSC. We employed two simulated datasets to train the model: one representing generated dynamical systems with randomly selected polynomial terms to model new disease behaviors, and another simulating noise-induced disease dynamics to account for noisy measurements. The model's performance was analyzed using both simulated data from different disease models and real-world data, including influenza and COVID-19. Results demonstrate that the proposed model outperforms previous models, effectively providing EWSs of impending outbreaks across various scenarios. This study bridges advancements in deep learning with the ability to provide robust early warning signals in noisy environments, making it highly applicable to real-world crises involving emerging disease outbreaks.
\end{abstract}


\section{Introduction}

Disease outbreaks caused significant burdens in the past decades, not only in terms of the global death counts but also by triggering economic crisis \cite{baker2022infectious, sweileh2022global, southall2021early}. Having an early warning helps with controlling the impact of the outbreak by applying preventive measures \cite{yang2017early, guo2020early, languon2019filovirus}. Statistical indicators, such as variance and lag-1 autocorrelation, are widely used for early warning signals (EWSs) which are known as generic Early Warning Indicators (EWIs) \cite{proverbio2022performance, southall2020prospects}. However, their performance varies with the type of noise involved in a system and they struggle to capture behaviors beyond the tipping point \cite{titus2020critical}. Additionally, these indicators typically produce signals only near the state change, making earlier detection challenging \cite{bury2021deep}.


From a machine learning perspective, generating an EWS by processing the outputs of a dynamical system, such as the number of infected cases, is a Time Series Classification (TSC) task. Machine learning models have achieved state-of-the-art performance in TSC and other fields \cite{foumani2023deep}.  Several models have been used to predict EWS for dynamical systems \cite{bury2021deep, deb2022machine, brett2020dynamical}. However, the predictive power of these models comes at the cost of requiring large datasets, which are often unavailable for emerging diseases. Furthermore, using historical data from past disease outbreaks also might result in a model that does not generalize well to new diseases.

One approach to address this challenge is to simulate training data using known dynamic models \cite{bury2021deep, deb2022machine}. Bury et al. \cite{bury2021deep} simulated training data using a two-dimensional dynamical model, where they randomly selected parameters to create four types of bifurcations: fold, Hopf, transcritical, and null (no bifurcation). They trained an LSTM-CNN model on this simulated data and tested it with simulated data generated from other known dynamical systems as well as empirical datasets. While the model outperformed generic EWS, it only considered one type of noise in the training data. Additionally, both the model and the generic EWS performed poorly on simulation data representing infectious disease spread.


Chakraborty et al. \cite{chakraborty2024early} generated a noise-induced Susceptible-Infected-Recovered (SIR) simulation dataset and trained the Bury et al. \cite{bury2021deep} LSTM-CNN model on this new data. The resulting model outperformed that original model \cite{bury2021deep} when tested against noise-induced disease-spreading models. However, the assessment focused on a single deep learning model, which had been fine-tuned using a different dataset \cite{bury2021deep}. Additionally, fixed-length time series were used as input, while real-world scenarios often involve time series of varying lengths. Furthermore, the model was trained, then evaluated, exclusively on noise-induced SIR simulation data, which may lead to overfitting.

Our goal is to develop and train a deep learning model that achieves higher accuracy than the previously mentioned models \cite{chakraborty2024early, bury2021deep} when faced with new, unforeseen disease outbreak data. We tested the best-performing machine learning models based on TSC benchmarks, fine-tuning and training the selected model using datasets from both Bury et al. \cite{bury2021deep} and Chakraborty et al. \cite{chakraborty2024early}. Our model forecasts transcritical bifurcations, enabling it to predict disease outbreaks in noisy environments. It can handle time series of variable lengths and generalizes effectively when confronted with unknown datasets.

\section{Methods} \label{model_formultion}
\subsection{Data}
The time series in this study represents the daily number of new infected cases recorded over time, providing a 1D sequence where each value corresponds to the count of newly identified cases for each day.

We used two datasets for training our models. (1) Bury et al. \cite{bury2021deep} simulated 200,000 time-series instances, using a two-dimensional dynamical system with polynomial terms that under certain conditions could exhibit three bifurcations: fold, Hopf, or transcritical. They randomly selected the parameters of the dynamical systems to include wide possible dynamics that might occur in unseen data. After finding the time series with an equilibrium point, they tweaked the parameters in a certain range to find possible bifurcations. Upon finding the bifurcation point, they simulated a null and a forced stochastic simulation. For the null case, they fixed the parameters while for the forced one, they linearly increased the parameter that caused the bifurcation from its equilibrium value to the bifurcation point while adding white noise to account for the stochasticity of real-world scenarios. They kept the last 1,500 time points before the bifurcation point, and repeated this process until having 50,000 time series for each type of bifurcation. In this study, we are interested only in the transcritical form of bifurcation that commonly appear in disease outbreaks. Hence, we only kept the transcritical and null parts of this dataset and we refer to this dataset as the RAPO dataset (Table \ref{dataset-table}).

(2) The Susceptible-Infected-Recovered (SIR) model is a common epidemiological model for explaining infectious disease dynamics \cite{brauer2008mathematical}. Chakraborty et al. \cite{chakraborty2024early} used the SIR model to generate new simulation data, aiming to train a better model than \cite{bury2021deep} for predicting impending transition in a disease outbreak, similar to predicting bifurcations in a dynamical system. They incorporated additive white noise, multiplicative environmental noise, and demographic noise in the SIR model to simulate the effect of stochasticity in real-world outbreaks. A total of 30,000 time series were simulated across the three noise-induced SIR models, with half containing transcritical bifurcations and the other half containing null bifurcations. Key parameters, such as disease transmission rates and noise intensity, were randomized during data generation to introduce variability and reflect uncertainty. In this study, we only used only the pre-transition portion of this dataset, which we refer to as the NISIR dataset (Table \ref{dataset-table}).




\begin{table}[hbt]
    \centering
    \begin{tabular}{|m{1 cm}||m{2 cm}|m{2 cm}|m{2 cm}|m{2 cm}|m{2 cm}|m{2 cm}|}
    \hline
    \rowcolor{gray!25}
         Name &Source&  Simulation model&  Bifurcation types &Simulation length &Noise component& Total number of simulations\\
    \hline
         NISIR & Chakraborty et al. \cite{chakraborty2024early} & SIR \newline + \newline noise & Trans \newline Null & 1,500 & Env \newline Dem \newline White & 90,000\\
    \hline
         RAPO & Bury et al. \cite{bury2021deep} & Two-dim dynamical systems with polynomial terms & Trans \newline Hopf \newline Fold \newline Null & 1,500 & White & 200,000\\
    \hline
    \end{tabular}
    \caption{Details of datasets used for training. \textbf{Trans})critical, \textbf{Hopf}, \textbf{Fold}, and \textbf{Null} are different bifurcations that are simulated in the data where Null means no bifurcation. (\textbf{Env})ironmental, (\textbf{Dem})ographic, and \textbf{White} noise were added to the simulated data.}
    \label{dataset-table}
\end{table}

Bury et al. \cite{bury2021deep} employed the CNN-LSTM architecture and tuned the model hyperparameters using their dataset. We adopted the same model architecture and hyperparameters. The reason we did not tune the hyperparameters with our dataset is that these hyperparameters include architecture-level parameters, and tuning them would result in a different model, whereas we want to compare our new models with that earlier one \cite{bury2021deep}, whose CNN layers first extract patterns from the input time series. The following LSTM layers extract temporal information from extracted patterns for classification, which is why we refer to it as seq-CNN-LSTM. Additionally, we tested another architecture that contains the LSTM and CNN layers, which implemented the LSTM and CNN layers in parallel, which we call par-CNN-LSTM. Our third model combined a one-dimensional Convolutional Layer with a self-attention mechanism, specifically using the Squeeze-and-Excite (SE) block \cite{hu2018squeeze}. This block offers channel-wise self-attention scores for CNN layers. We referred to it as Conv1d+SE. Finally, we evaluated other top-performing models in TSC benchmarks \cite{foumani2023deep} such as InceptionTime \cite{ismail2020inceptiontime} and SAnD, a transformer-based Time Series Classifier \cite{song2018attend}.

\subsection{Training and testing}
We preprocessed the raw dataset following the same procedure as in \cite{bury2021deep}, which censored data instances from both the beginning and end of the time series. We did this because the time before a disease outbreak is unknown (end) and prevalence data is often not recorded at early stages (beginning). If a time series is 1500 points long, we selected two random numbers from the interval [0, 725], then removed ending chunks of the time series by making them zero, for censoring each time series. After censoring, we normalized each instance with its average. Furthermore, we detrended the time series with a Lowess smoothing \cite{cleveland1979robust}.

We created a dataset comprising a total of 160,000 time series from two sources.  We use the Python \texttt{scikit-learn} package to split the preprocessed dataset into three sets using an 80/15/5 ratio for training, validation, and testing, using stratified sampling to ensure each set has an equal number of bifurcations. Using five percent of the data for testing results in 8,000 time series instances which is big enough to test the model's performance. 

Our study was conducted in three steps. In the first step, we trained each of the three model architectures on each of the three noise-induced SIR simulated datasets, resulting in a total of nine models. Before training, we fine-tuned the hyperparameters of the models except for the seq-CNN-LSTM model, for which we used the hyperparameters from \cite{bury2021deep}. We used the \texttt{kerastuner} library's implementation of Bayesian Optimization (BO) with Gaussian Processes to select the hyperparameters \cite{osborne2009gaussian}, including model architecture such as the number of CNN layers, number of filters per CNN layer, number of LSTM layers, number of memory cells per LSTM layers, and convergence hyperparameters such as learning rate, dropout rate, kernel regularizer l2 coefficient, a regularizer that penalizes layer weights. We used the training and validation set to select the best combination of hyperparameters. To prevent overfitting, we applied \emph{early stopping} \cite{prechelt2002early}.

In the second step, we selected the best-performing model from the first step. We then combined the three noise-induced SIR simulated datasets and re-performed the hyperparameter tuning and training. In the third step, we added the RAPO dataset to our combined dataset and repeated the hyperparameter tuning and training process.

We tested our model's capability of generalizing in three levels. At each level, it is harder for the model to maintain a good performance. In the first step, we used the three noise-induced test sets and calculated accuracy and f1-score. To compare our model with the model of \cite{bury2021deep}, we additionally tested all models with the RAPO test set. Since Chakraborty et al.'s model was trained on uncensored data, we performed a fair comparison by also testing all models on the uncensored versions of all four datasets.
In the second step, we tested our model with data simulated from two compartmental models in epidemiology that exhibits transcritical bifurcations. One of them is Susceptible-Exposed-Infectious–Recovered (SEIR) model (SI Appendix), and another one is Susceptible-Exposed-Infectious-Removed-Vaccinator (SEIRx) model (SI Appendix). For testing, we followed the same procedure as in Bury et al. \cite{bury2021deep}. For each model, we generated 20 time sequences with half of them for each class: transcritical and null. For classification, we considered the last 20\% of each instance. We evaluated model performance using Area-under the Receiver Operator Characteristics (ROC) curve (AUC) for these tests \cite{spackman1989signal}. The ROC curve illustrates the trade-off between true positive and false positive (i.e., transcritical and null bifurcations). A perfect classifier, which predicts all cases correctly, yields an AUC of 1, while a random classifier yields an AUC of 0.5.

In the third step, we tested our model using influenza data from the United Kingdom and COVID-19 data from Edmonton. The COVID-19 data was obtained from the City of Edmonton's Open Data Portal \cite{COVID-19Edmonton} and the influenza data was sourced from Our World in Data \cite{owid-influenza}. The data was preprocessed using the same methodology as Chakraborty et al. \cite{chakraborty2024early}. This test was particularly significant, as it evaluated the model's ability to generalize from simulation datasets to real-world scenarios.

\section{Results}
The InceptionTime model did not perform as well as the other models. Similarly, the transformer-based model required extensive hyperparameter tuning to achieve reasonable results, but its performance remained suboptimal. Consequently, we did not move forward with these two models. Instead, we narrowed our focus to three models to limit our search space: par-LSTM-CNN, seq-LSTM-CNN, and Conv1d+SE.

All three models performed almost equally well against the NISIR test sets, with par-LSTM-CNN outperforming seq-LSTM-CNN (using the hyperparameters from Bury et al. \cite{bury2021deep}) and the Conv1d+self-attention models (Table \ref{tab:1}). Hence par-LSTM-CNN was selected as the best model for further analysis. Trained on the three noise-induced datasets, par-LSTM-CNN matched the performance of the best models trained individually on each noise-induced dataset. However, its performance on the RAPO test set was only marginally better than chance, achieving an improvement of just five percent (Table \ref{tab:2}).

The par-LSTM-CNN model, trained on the NISIR and RAPO datasets and referred to as "our model," demonstrated comparable performance to Chakraborty et al.'s model on the NISIR datasets and to Bury et al.'s model on the RAPO test set. However, Chakraborty et al.'s model performed poorly on the RAPO test set, while Bury et al.'s model exhibited low accuracy on the NISIR test set (Table \ref{tab:3}). Notably, our model achieved the highest accuracy when tested with censored instances (Table \ref{tab:4}). Against censored NISIR test sets, only our model maintained strong performance. In contrast, Bury et al.'s model failed to maintain accuracy when faced with unseen data from the NISIR dataset, and Chakraborty et al.'s model struggled to handle censored data. Furthermore, on the RAPO test set, our model surpassed Bury et al.'s model in performance, while Chakraborty et al.'s model continued to underperform (Table \ref{tab:4}).

\begin{table}[htb!]
    \centering
    \begin{tabular}{|m{3cm}|m{3cm}|m{3cm}|m{3cm}|}
        \hline
         Model & Part of NISIR dataset used & Validation Accuracy & Test Accuracy \\
        \hline
         seq-LSTM-CNN & White & 0.9907 & \\
        \hline
         CONV1d+SE & White & 0.9968 & \\
        \hline
         par-LSTM-CNN & White & \textbf{0.997} & 0.996 \\
        \hhline{|=|=|=|=|}
         seq-LSTM-CNN & Env & 0.9987 & \\
        \hline
         CONV1d+SE & Env & 0.9985 & \\
        \hline
         par-LSTM-CNN & Env & \textbf{0.9997} & 0.998 \\
        \hhline{|=|=|=|=|}
         seq-LSTM-CNN & Dem & 0.998 & \\
        \hline
         CONV1d+SE & Dem & 0.9976 & \\
        \hline
         par-LSTM-CNN & Dem & \textbf{0.9987} & 0.999 \\
        \hline
    \end{tabular}
    \caption{Comparing the performance of the three model architectures using the validation set accuracy and test set f1-score against three noise-induced datasets, namely White Noise (White), Environmental Noise (Env), and, Demographic Noise (Dem). We used validation accuracy to select the best model and tested the out-of-sample accuracy using the test set from each part of the NISIR dataset.}
    \label{tab:1}
\end{table}

\begin{table}[htb!]
    \centering
    \begin{tabular}{|m{2 cm}||m{1.5 cm}|m{1.5 cm}|m{2 cm}|m{2 cm}|m{2 cm}|m{2 cm}|}
        \hline
         Model & Training Dataset & Val Acc & Env Test Acc & Dem Test Acc & White Test Acc & RAPO Test Acc \\
        \hline
         par-LSTM-CNN & NISIR & 0.9976 & 0.998 & 0.999 & 0.997 & 0.557 \\
        \hline
    \end{tabular}
    \caption{Validation set (Val) and test accuracy of the par-LSTM-CNN model trained with all parts of the NISIR dataset against four test sets: the three test sets from the NISIR dataset and the RAPO test set.}
    \label{tab:2}
\end{table}

\begin{table}[htb!]
    \centering
    \begin{tabular}{|m{2.5cm}|m{2cm}|m{2cm}|m{2cm}|m{2cm}|m{2cm}|}
        \hline
         Model & Training Dataset & Env Test Acc & Dem Test Acc & White Test Acc & RAPO Test Acc \\
        \hline
         par-LSTM-CNN & NISIR and RAPO & 0.996 & 0.998 & 0.987 & 0.500 \\
        \hline
         Bury et al.'s model & RAPO & 0.752 & 0.834 & 0.649 & 0.486 \\
        \hline
         Chakraborty et al.'s model & NISIR & 0.990 & 0.995 & 0.967 & 0.499 \\
        \hline
    \end{tabular}
    \caption{Comparing the accuracies (Acc) of Bury et al.'s model and Chakraborty et al.'s model with our best model, par-LSTM-CNN tuned and trained with \textbf{Uncensored} versions of NISIR i.e., White noise (White), Demographic noise (Dem), and Environmental noise (Env), and RAPO test set.}
    \label{tab:3}
\end{table}

\begin{table}[htb!]
    \centering
    \begin{tabular}{|m{2.5cm}|m{2cm}|m{2cm}|m{2cm}|m{2cm}|m{2cm}|}
        \hline
         Model & Training Dataset & Env Test Acc & Dem Test Acc & White Test Acc & Bury et al. Test Acc \\
        \hline
         par-LSTM-CNN & NISIR and RAPO & 0.995 & 0.996 & 0.986 & 0.817 \\
        \hline
         Bury et al.'s model & RAPO & 0.682 & 0.629 & 0.559 & 0.747 \\
        \hline
         Chakraborty et al.'s model & NISIR & 0.536 & 0.545 & 0.562 & 0.511 \\
        \hline
    \end{tabular}
    \caption{Comparing the accuracies (Acc) of Bury et al.'s model and Chakraborty et al.'s model with our best model, par-LSTM-CNN tuned and trained with \textbf{Censored} versions of NISIR i.e., White noise (White), Demographic noise (Dem), and Environmental noise (Env), and RAPO test set.}
    \label{tab:4}
\end{table}

\begin{figure}[htp!]
    \centering
    \begin{subfigure}{0.49\textwidth}
        \includegraphics[width=\linewidth]{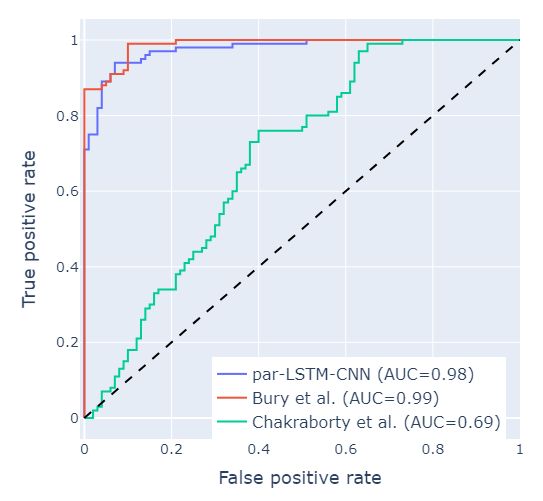}
        \put(-215,185){\textbf{(a)}}
        \label{fig:image1}
    \end{subfigure}
    \hfill
    \begin{subfigure}{0.49\textwidth}
        \includegraphics[width=\linewidth]{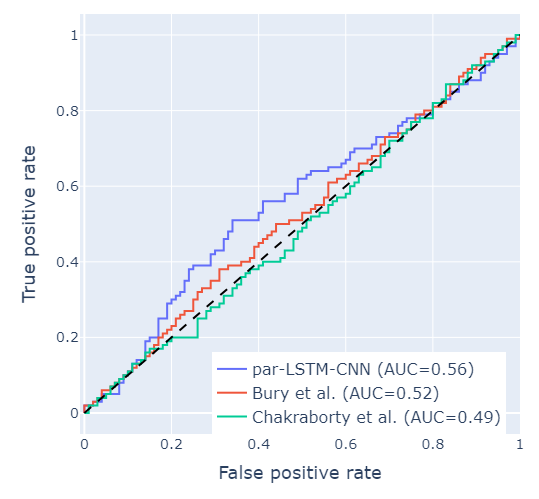} 
        \put(-215,185){\textbf{(b)}}
        \label{fig:image2}
    \end{subfigure}

    \vspace{-0.05cm}

    \begin{subfigure}{0.49\textwidth}
        \includegraphics[width=\linewidth]{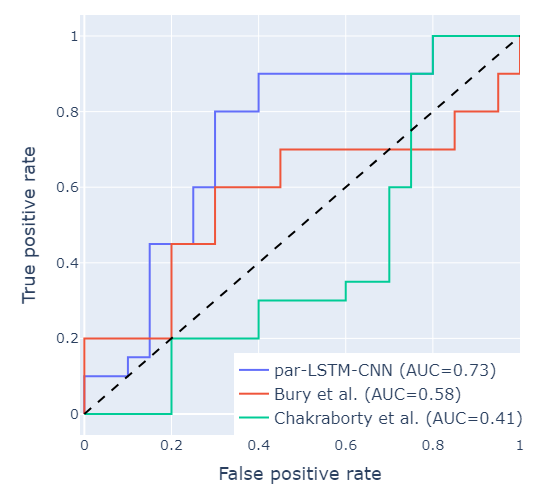} 
        \put(-215,185){\textbf{(c)}}
        \label{fig:image3}
    \end{subfigure}
    \hfill
    \begin{subfigure}{0.49\textwidth}
        \includegraphics[width=\linewidth]{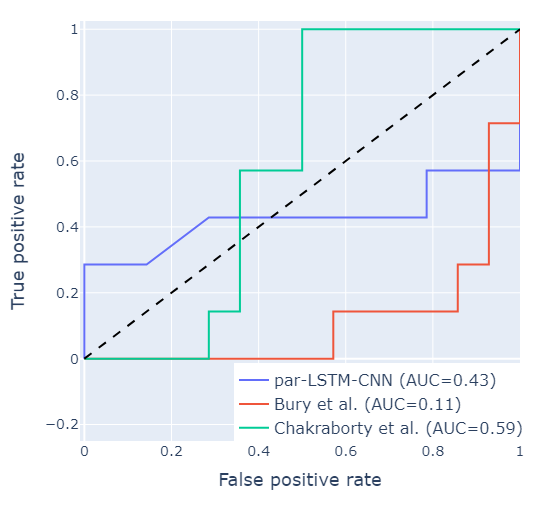} 
        \put(-215,185){\textbf{(d)}}
        \label{fig:image4}
    \end{subfigure}

    \caption{Area under Receiver Operator Characteristics curve (AUC) of our model (par-LSTM-CNN), Bury et al.'s model, and Chakraborty et al.'s model in tests with (A) SEIR model, (B) SEIRX model, (C) influenza data, and (D) COVID-19 data.}
    \label{ROC_figure}
\end{figure}

In testing against the SEIR dataset, our model performed comparably to Bury et al.'s model, achieving an AUC of approximately 0.99 (Figure \ref{ROC_figure}). In contrast, Chakraborty et al.'s model showed significantly lower performance, with an AUC of 0.69 for the same dataset (Figure \ref{ROC_figure}). Against the number of infected (I) variable of the SEIRx model, all three models performed poorly, although our model demonstrated a slight performance advantage (Figure \ref{ROC_figure}).

For the influenza dataset, our model outperformed the other two, achieving an AUC of 0.73 (Figure \ref{ROC_figure}). Chakraborty et al.'s model exhibited the lowest performance, with an AUC of 0.41. Conversely, for the COVID-19 dataset, Chakraborty et al.'s model surpassed the other two, achieving an AUC of 0.59 (Figure \ref{ROC_figure}). Bury et al.'s model showed the weakest performance, with an AUC of 0.11. Our model also performed poorly, but better than Bury et al., on this dataset, with an AUC of 0.43 (Figure \ref{ROC_figure}).


\newpage

\section{Discussion} \label{conclusion}

We trained deep learning models to predict disease outbreaks using two simulation datasets. The first dataset was the noise-induced SIR (NISIR) model simulation data \cite{chakraborty2024early}, which incorporates both the dynamics of infectious diseases and the stochasticity present in real-world datasets. The second dataset was generated using randomly selected polynomials (RAPO) \cite{bury2021deep}. To avoid overfitting and ensure generalization to potentially unknown diseases with dynamics differing from those described by the SIR model, we used the transcritical portion of the RAPO simulations. We evaluated the best-performing deep learning models from the Time Series Classification benchmarks \cite{foumani2023deep} and selected the optimal model based on the AUC metric. To assess the performance improvements of our approach, we compared our models to the pre-trained models described in \cite{chakraborty2024early, bury2021deep} using simulation models of infectious diseases. To determine whether predicting bifurcations in simulation data could translate to predicting disease outbreaks in real-world data, we tested our model on empirical datasets for influenza and COVID-19. Notably, our model outperformed other models on the influenza dataset and demonstrated equal performance on simulated datasets.

All deep learning models achieved near-perfect scores when trained and tested on the NISIR dataset. This indicates that these models are more complex than the dataset itself, suggesting they would perform well on datasets with a higher level of induced noise. This trend persisted when the three noise-induced parts of the NISIR dataset were combined. The par-LSTM-CNN model, which performed slightly better than the others when trained on each of the three parts individually, achieved the same accuracy when trained on the entire NISIR dataset. Increasing the size of the dataset improved the generalization capability of this deep learning model in handling various types of noise. However, the model failed to generalize to an out-of-sample test set from the RAPO dataset. Features learned by the model from the NISIR dataset explained the dynamics of the RAPO dataset only 5\% better than chance. In contrast, features learned by Bury et al.'s model from the RAPO dataset explained the NISIR dataset up to 18\% better than chance. This aligns with our expectations, as the RAPO dataset exhibits greater diversity by generating dynamical systems with randomly selected polynomial terms.

A model that achieves high accuracy with a censored setup is expected to perform well when more data is available in an uncensored setup. The performance difference between Chakraborty et al.'s model and the other two models lies in the training data: Chakraborty et al.'s model was trained using pre-transition data without censoring. However, all models performed no better than chance when tested on the uncensored RAPO dataset. It is important to note that Bury et al.'s sub-optimal accuracy on their RAPO dataset does not directly reflect their model's performance in the final tests, as these tests involved nearly uncensored data instances. Nevertheless, our model outperformed Bury et al.'s model by 5\% in a fair comparison using a test set from the RAPO dataset that both models had utilized during training. The performance improvement of our model, which was trained using both the RAPO and NISIR datasets, is attributable to two factors: the use of more data and the optimization of the model architecture through Bayesian Optimization. This architectural optimization enabled our model to achieve 1.5\% better accuracy than Chakraborty et al.'s model on the NISIR test set and 5\% better accuracy on the RAPO test set, even when using the same training data. Additionally, the enhanced generalization capability of our model, trained with the more diverse RAPO and NISIR datasets, contributed significantly to its superior performance compared to the other two models.

In tests with the SEIR and SEIRx models, our model demonstrated slightly better performance in one scenario and comparable performance in others. The consistently poor results across all models for variable I of the SEIRx model underscore the substantial differences in dynamics between this model and the training datasets. When tested on real-world datasets, our model outperformed the other models on the influenza dataset but showed lower performance on the COVID-19 dataset. This lower performance could be attributed to the shorter length of the real-world data, particularly the presence of null time series. 


Our study has several limitations. From a deep learning perspective, we evaluated several architectures, including CNN, LSTM, Attention, and Transformers. Ultimately, we selected the LSTM-CNN model, as the other architectures did not perform as well. Although the Transformer-based architecture did not yield strong results with Bayesian optimization in this study, its superior performance in recent time series classification (TSC) benchmarks \cite{foumani2023deep} suggests that it may be a promising candidate for this type of data. With sufficient training and hyperparameter tuning, Transformer-based models could potentially provide a more robust deep learning solution.

Additionally, while our model was trained on long time series consisting of 1500 data points, real-world scenarios—such as the emergence of a new disease—may not provide such extensive records. To better align with real-world conditions, it would be beneficial to train models using shorter time series. Another potential approach involves training models on time series of varying lengths and combining their predictions through an ensemble method. From the perspective of dynamical systems, the SEIRx test results underscore the need for greater diversity in the training set. For instance, simulating SEIRx with different noise sources could introduce an additional layer of complexity and realism. A further limitation concerns the simulation of the bifurcation parameter's behavior. In both datasets used, this parameter increased linearly until the shifting point; however, real-world scenarios may exhibit more intricate patterns \cite{chakraborty2024policy}, which our current model does not account for.

While this study has several limitations,  it is a step forward for using deep learning advancement and dynamical systems to help control the growing danger of infectious diseases.

\section*{Data accessibility}
The training and testing data used in this study were collected from Bury et al. \cite{bury2021deep} and Chakraborty et al. \cite{chakraborty2024early}. Influenza data was collected from Our World in Data \cite{owid-influenza} and COVID-19 data was collected from \cite{COVID-19Edmonton}. 

\section*{Funding}
This project was primarily supported by One Health Modelling Network for Emerging Infections (OMNI). Hao Wang's research was partially supported by the Natural Sciences and Engineering Research Council of Canada (Individual Discovery Grant RGPIN-2020-03911 and Discovery Accelerator Supplement Award RGPAS-2020-00090) and the Canada Research Chairs program (Tier 1 Canada Research Chair Award). Mark A. Lewis gratefully acknowledges support from an NSERC Discovery Grant and the Gilbert and Betty Kennedy Chair in Mathematical Biology. Pouria Ramazi also acknowledges the support from an NSERC Discovery Grant. Russell Greiner was also partially funded by CIFAR and NSERC. 

\section*{Ethics}
This work did not require ethical approval from a human subject or animal welfare committee.

\section*{Declaration of AI use}
We have not used AI-assisted technologies in creating this article.

\section*{Authors’ contributions}
R.M.: conceptualization, formal analysis, investigation, methodology, software, validation, visualization, writing-original draft; A.K.C: software, validation, visualization, writing-original draft; R.G.: conceptualization, funding acquisition, methodology, and editing; M.A.L.: conceptualization, funding acquisition, project administration, supervision; H.W.: conceptualization, funding acquisition, methodology, project administration, supervision; T.G.: conceptualization, methodology, and editing; P.R.: conceptualization, funding acquisition, methodology, and editing. 

\section*{Conflict of interest declaration} We declare we have no competing interests.

\bibliographystyle{unsrt}
\bibliography{MS}

\appendix
\section{SI Appendix}

\subsection{SEIR model}
The susceptible–exposed-infectious–recovered (SEIR) model equations \cite{brauer2008mathematical} are given by 
\begin{equation}
\begin{cases}
\frac{d S(t)}{d t} \;=\; \Lambda \,-\, \beta(t) S(t) I(t) \,-\, d S(t) \,+\, \sigma_1 dW_1(t), \\
\frac{d E(t)}{d t} \;=\; \beta(t) S(t) I(t) \,-\, (d+\kappa) E(t) 
,+\, \sigma_2 dW_2(t), \\
\frac{d I(t)}{d t} \;=\; \kappa E(t) \,-\, (d+\gamma) I(t) \,+\, \sigma_3 dW_3(t), , \\
\frac{d R(t)}{d t} \;=\; \gamma I(t) +\, \sigma_4 dW_4(t) \,-\, d R(t) \,+\, \sigma_4 dW_4(t), 
\end{cases}
\label{SEIR_model}
\end{equation}
where $S(t)$, $E(t)$, $I(t)$, and $R(t)$ denote the susceptible, exposed, infectious, and recovered individuals, respectively, at time $t$. Here, $\Lambda$ is the recruitment rate of the susceptible population, $\beta(t)$ is the disease transmission rate, $d$ is the natural death rate, $1/ \gamma$ is the mean infectious period, and $1/ \kappa$ is the mean exposed period. $\sigma_i,$ $i = 1, 2, 3, 4,$  are the intensities of white noise and $W_i(t),$ $i = 1, 2, 3, 4,$ are independent Wiener processes. The basic reproduction number of the model is $\frac{\kappa \beta \Lambda}{d (d \,+\, \kappa) (d \,+\, \gamma)}$, and the bifurcation point is $\beta_c = \frac{d (d \,+\, \kappa) (d \,+\, \gamma)} {\kappa \Lambda}$. 



\subsection{SEIRx model}

The stochastic version of Susceptible-Exposed-Infectious-Removed-vaccinator (SEIRx) model equation \cite{bury2021deep} are given by
\begin{align}
\begin{cases}
\frac{dS}{dt} &= \mu N (1 - x) - \mu S - \beta \frac{SI}{N} + \sigma_1 \xi_1(t), \\
\frac{dE}{dt} &= \beta \frac{SI}{N} - (\sigma + \mu) E + \sigma_2 \xi_2(t), \\
\frac{dI}{dt} &= \sigma E - (\gamma + \mu) I + \sigma_3 \xi_3(t), \\
\frac{dR}{dt} &= \mu x + \gamma I - \mu R + \sigma_4 \xi_4(t), \\
\frac{dx}{dt} &= \kappa x (1 - x) \left( - \omega + I + \delta(2x - 1) \right) + \sigma_5 \xi_5(t),
\end{cases}
\end{align}
where \( S \) represents the population of individuals who are susceptible to the disease, \( E \) refers to those who have been exposed to the infection but are not yet infectious, \( I \) denotes the individuals who are infectious, and \( R \) is the group of individuals who have recovered or developed immunity. The variable \( x \) represents individuals with a provaccine sentiment. The parameters include \( \mu \), the per capita birth and death rate, \( \beta \), the transmission rate of the disease, \( \sigma \), the per capita rate at which exposed individuals become infectious, \( \gamma \) is the recovery rate from the infection, \( \kappa \) is the rate of social learning, \( \delta \) indicates the strength of injunctive social norms, and \( \omega \) reflects is the perceived relative risk of vaccination versus infection, $\sigma_i,$ $i = 1, 2, 3, 4, 5,$ are the noise amplitudes, and $\xi_i,$ $i = 1, 2, 3, 4, 5,$ are Gaussian white noise process.

\end{document}